\documentclass[10pt,twocolumn,letterpaper]{article}

\usepackage{iccv}
\usepackage{times}
\usepackage{epsfig}
\usepackage{graphicx}
\usepackage{amsmath}
\usepackage{amssymb}

\usepackage{amsfonts}
\usepackage{bm}
\usepackage{subfigure}
\usepackage{booktabs}
\usepackage{tabu}
\usepackage{multirow}
\usepackage{color}
\usepackage[abs]{overpic}
\usepackage{comment}

\def\etc{\emph{etc.}}

\usepackage[breaklinks=true,bookmarks=false]{hyperref}

\iccvfinalcopy 

\ificcvfinal\fi

\newcommand{\mymodel}{Saliency-Associated Object Tracker\xspace}

\newcommand*{\affaddr}[1]{#1} 
\newcommand*{\affmark}[1][*]{\textsuperscript{#1}}
\newcommand*{\email}[1]{\small{\texttt{#1}}}

\begin{document}

\title{Saliency-Associated Object Tracking}

\author{%
Zikun Zhou\affmark[1], Wenjie Pei\affmark[1,*], Xin Li\affmark[2], Hongpeng Wang\affmark[1,2], Feng Zheng\affmark[3], and Zhenyu He\affmark[1,*]\\
\affaddr{\affmark[1]Harbin Institute of Technology, Shenzhen}\quad
\affaddr{\affmark[2]Peng Cheng Laboratory}\\
\affaddr{\affmark[3]Southern University of Science and Technology}\\
\email{zhouzikunhit@gmail.com\quad wenjiecoder@outlook.com\quad xinlihitsz@gmail.com}\\ \email{wanghp@hit.edu.cn\quad zfeng02@gmail.com\quad zhenyuhe@hit.edu.cn}\\}

\maketitle
\renewcommand{\thefootnote}{\fnsymbol{footnote}} 
\footnotetext[1]{Corresponding authors.}

\ificcvfinal\thispagestyle{empty}\fi

\begin{abstract}
Most existing trackers based on deep learning perform tracking in a holistic strategy, which aims to learn deep representations of the whole target for localizing the target. It is arduous for such methods to track targets with various appearance variations.
To address this limitation, another type of methods adopts a part-based tracking strategy which divides the target into equal patches and tracks all these patches in parallel.
The target state is inferred by summarizing the tracking results of these patches. 
A potential limitation of such trackers is that not all patches are equally informative for tracking. Some patches that are not discriminative may have adverse effects.
In this paper, we propose to track the salient local parts of the target that are discriminative for tracking. In particular, we propose a fine-grained saliency mining module to capture the local saliencies.
Further, we design a saliency-association modeling module to associate the captured saliencies together to learn effective correlation representations between the exemplar and the search image for state estimation.
Extensive experiments on five diverse datasets demonstrate that the proposed method performs favorably against state-of-the-art trackers.
\vspace{-5mm}
\end{abstract}

\section{Introduction}
\vspace{-1mm}
Visual object tracking aims to predict the target states in a tracking sequence given the initial state of the target object in the first sequence frame. It is a fundamental research topic in Computer Vision and has a wide range of applications including video surveillance, robotics, and motion analysis.
Although deep trackers~\cite{ASRCF,MDNet,CREST,SiamAttn}, which benefit from excellent feature learning for images by deep neural networks, have achieved great progress in recent years, tracking targets with various real-time appearance variations, such as deformation, occlusion, and viewpoint changes, \etc, remains an extremely challenging task.

\begin{figure}[t]
\centering
	\includegraphics[width=1.0\columnwidth]{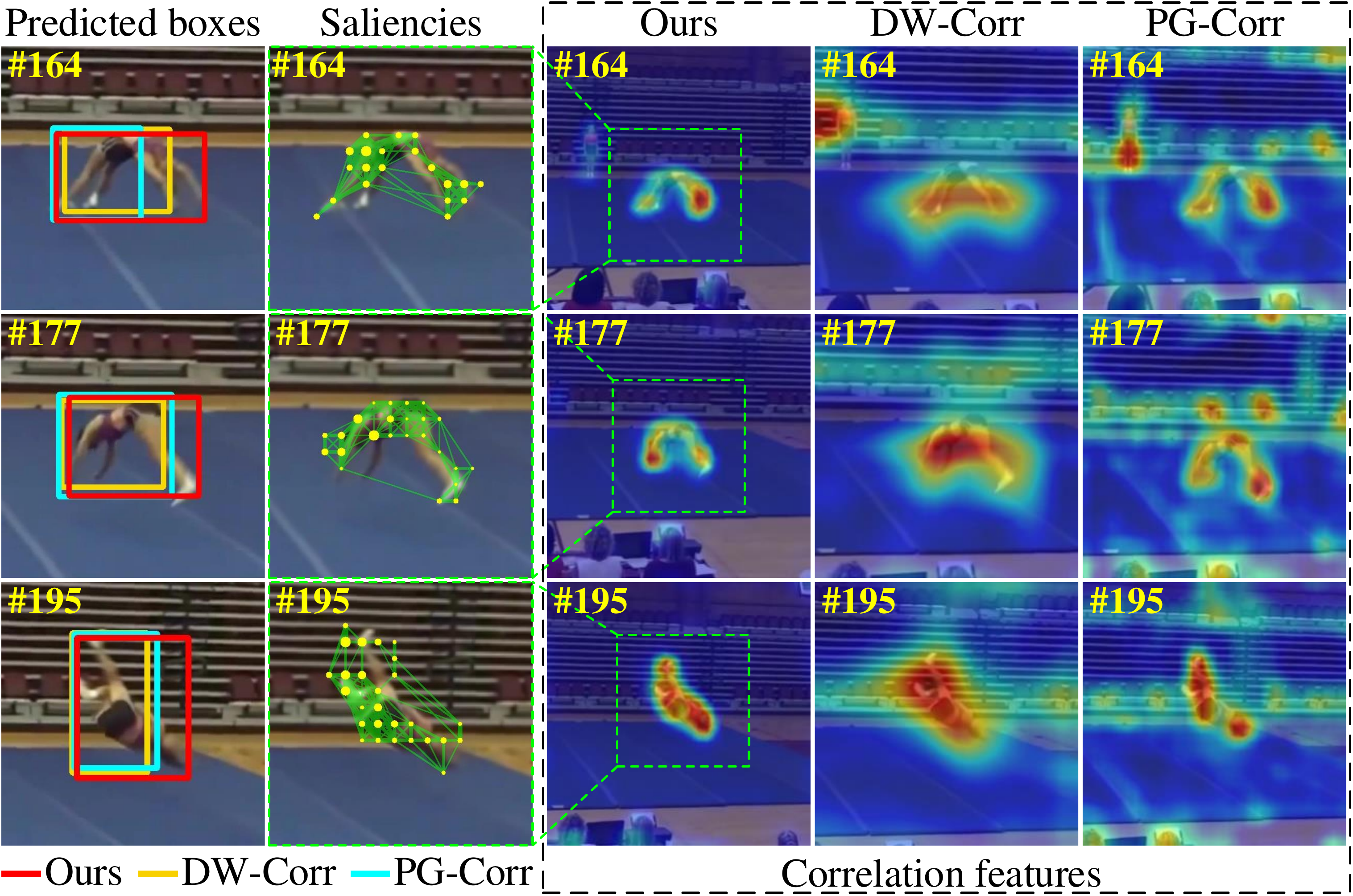}
	\caption{Given a search image in a tracking sequence, our \emph{SAOT} first captures local saliencies (yellow dots) of the target that are discriminative for tracking, and then associates the captured saliencies together to learn precise correlations between the target exemplar and the search image for reflecting target states. Thus, our model can generate more precise correlation features than DW-Corr~\cite{SiamRPN++} (in the holistic tracking strategy) and PG-Corr~\cite{PG-Net} (in the part-based strategy), and accordingly predict more precise bounding boxes. The correlation features are visualized by averaging all channels (red color indicates higher correlation). Larger-size salient dots indicate higher saliency values.}
	\vspace{-5mm}
	\label{Fig:Introduction}
\end{figure}

A classical type of deep tracking approaches~\cite{DiMP,ATOM,SiamRPN++,OCEAN} performs tracking in a holistic strategy, which seeks to learn a precise deep feature representation for the whole target object and then localize the target in the search image. A prominent example is the Siamese-based trackers~\cite{SiamFC,SiamRPN++,SiamRPN,TADT,SINT}, which learn deep representations for both the target exemplar and the search image in the same feature space by a Siamese neural network, and then perform target tracking by feature matching between them.
Such methods perform well in ordinary scenarios in which the target keeps stable appearances close to the exemplar, but struggle in the challenging scenarios where the target varies substantially.
This is because the global appearance gap between the target exemplar and the target state in the search image results in an inevitable tracking error. The online learning trackers~\cite{ECO,KCF,CREST}, which are another typical type of methods, are designed to adapt to the appearance variations of the target by learning an online filter. However, these methods still perform tracking in the holistic strategy and thus can hardly deal with drastic appearance variations.

In contrast to the holistic tracking strategy, another type of existing tracking methods~\cite{PG-Net,SCF,PMT,RSST} adopts the part-based strategy, which first tracks local parts of the target object and then infers the target state by summarizing the tracking results of these parts. A common way of these part-based methods is to partition the target into regular patches equally and then perform tracking on all these patches in parallel. Whilst such a part-based tracking strategy mitigates the difficulties of tracking appearance-varying targets, a potential limitation is that not all the partitioned patches are equally informative for tracking. Some parts which are not discriminative are difficult to be tracked and may have adverse effects on inferring the global target state.

In this paper, we follow the part-based tracking strategy and propose the Saliency-Associated Object Tracker (\emph{SAOT}). The key difference between our \emph{SAOT} and other part-based tracking methods is that \emph{SAOT} focuses on capturing and tracking the local saliencies of the target that are discriminative for tracking instead of simply tracking all partitioned patches in parallel. Specifically, we design a fine-grained saliency mining mechanism to capture local saliencies in the target that are discriminative and easily localized in the search image. Subsequently, these captured saliencies are associated together by modeling the interactions between them to learn global correlations between the target exemplar and the search image, which can reflect the target state in the search image precisely.

The rationales behind such design of our \emph{SAOT} are:
1) the salient local regions in the target, which are tracked more precisely and easily than other regions, can potentially keep consistent distinctiveness in various appearance variations;
2) different associations between the saliencies correspond to different appearances of the same target, so that we model the associations between the captured saliencies to adapt to real-time appearance variations. 
Consequently, our \emph{SAOT} is able to cope with various appearance variations of the target during tracking, such as deformation and occlusion. Figure~\ref{Fig:Introduction} presents an example of tracking a gymnast, in which the appearance of the gymnast varies substantially during display. 
Owing to the captured saliencies robust to appearance variations, the bounding boxes predicted by our model are much more precise than those predicted based on DW-Corr~\cite{SiamRPN++} and PG-Corr~\cite{PG-Net}, which are in the holistic strategy and the part-based strategy, respectively.

The tracking strategy of the proposed \emph{SAOT}, which first deals with local saliencies with high confidence and then associates them together to achieve the global solution, is akin to the divide-and-conquer algorithm. 
To conclude, we make the following contributions: 1) A fine-grained saliency mining module is designed to capture local saliencies in the target which are discriminative for tracking.
    2) We propose a saliency-association modeling module to associate the captured saliencies together to learn effective global correlations between the exemplar and the search image.
    3) We achieve favorable performance against state-of-the-art methods in both quantitative and qualitative evaluations on five benchmarks (OTB2015, NFS30, LaSOT, VOT2018, and GOT10k), demonstrating the effectiveness of our \emph{SAOT}.
\vspace{-6mm}

\section{Related Work}
\vspace{-1mm}
This section mainly discusses the related trackers from the perspectives of the holistic and part-based strategies.

\vspace{0.4mm}
\noindent\textbf{Holistic-strategy trackers.}
Numerous Siamese-based trackers~\cite{SiamFC,SiamRPN++,SiamRPN,SINT} perform tracking in the holistic strategy. Such trackers measure the similarity between the exemplar and the search image by feature matching to localize the target, in which the feature maps of the exemplar are treated as a holistic kernel to perform cross-correlation on the search image. Most of them~\cite{SiamFC, SiamRPN++,SiamRPN,OCEAN} use the target from the first frame as a fixed exemplar to track the target in all subsequent frames, resulting in limited robustness to appearance variations of the target during tracking. Several adaptive Siamese-based methods~\cite{GCT,Gradnet,UpdateNet,FlowT}, which use the historical target states to update the representation of the exemplar, are proposed to address this limitation.

Many online learning trackers~\cite{KCF,CREST} also perform tracking in the holistic strategy. These trackers learn a correlation filter~\cite{ASRCF,ECO,CCOT,KCF} or a convolutional filter~\cite{DiMP,ATOM,CREST} using online collected samples, and use the filter as a holistic appearance model to distinguish the target from backgrounds. Although the adaptive Siamese-based trackers~\cite{Gradnet,UpdateNet,FlowT} and online learning trackers~\cite{KCF,CREST} model the target information from historical frames, they are less effective in handling drastic real-time appearance variations of the target due to the holistic tracking strategy.

\begin{figure*}[t]
\centering
    \includegraphics[width=0.93\textwidth]{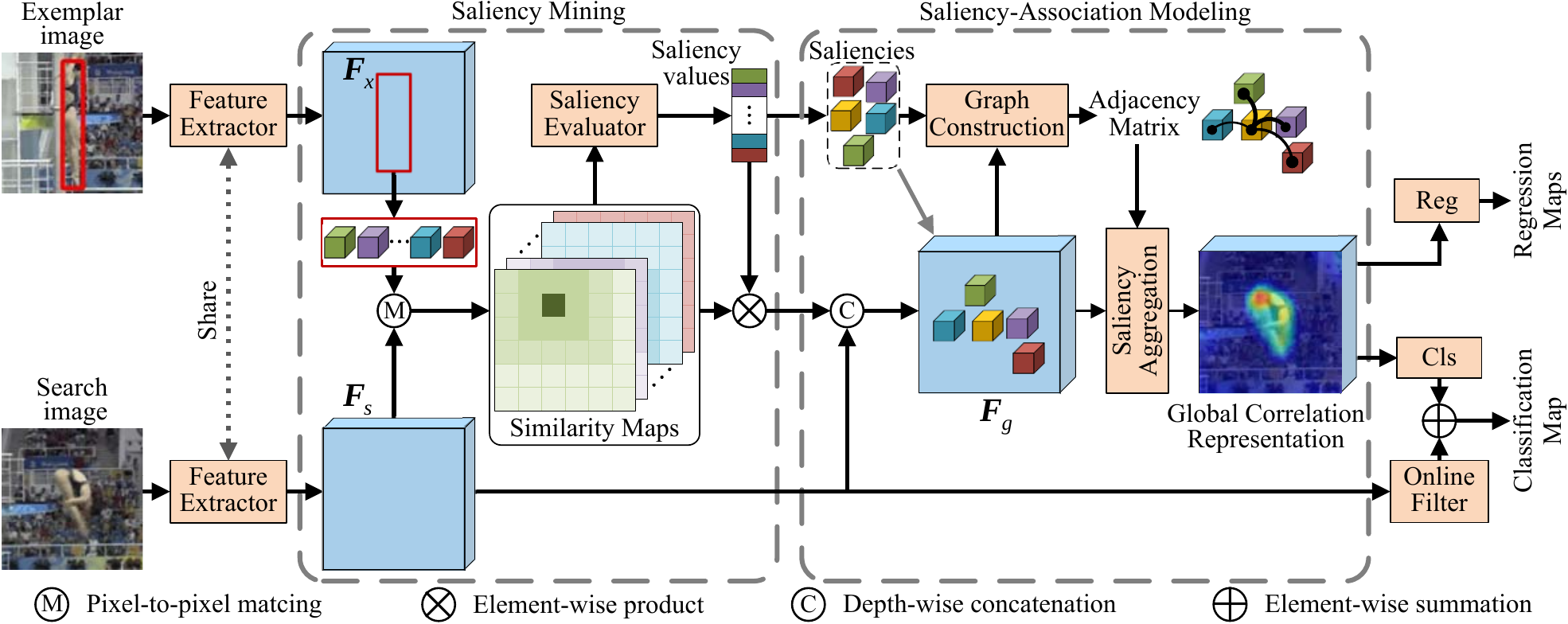}
    \vspace{-0.5mm}
    \caption{\textbf{Architecture of the proposed Saliency-Associated Object Tracker (\emph{SAOT}).} It contains two core modules: 1) Saliency Mining module, which captures the saliencies of the target; 2) Saliency-Association Modeling module, which associates the captured saliencies to learn an effective correlation representation for state estimation. Reg and Cls denote the regression and classification heads, respectively.}
\label{Fig:Framework}
\vspace{-4mm}
\end{figure*}

\vspace{0.4mm}
\noindent\textbf{Part-based trackers.}
Many traditional trackers~\cite{RPT,SCF,PMT,SST,RSST} resort to the part-based strategy to handle the challenges of deformation and occlusion. 
Most of them~\cite{SCF,PMT,SST,RSST} directly track all the equally-partitioned patches of the target in parallel, instead of selecting the patches easy to be tracked according to their discriminability.
As a result, the less discriminative patches may adversely affect the adaptability of these approaches.
RPT~\cite{RPT} estimates the reliability for the randomly sampled patches of the target in a Monte Carlo framework and tracks reliable patches with multiple traditional correlation filters.
However, the predicted positions of the patches are combined using a voting scheme in RPT, which can only estimate a coarse target state.
In addition, the above part-based trackers are designed based on the less representative hand-crafted features, which limits their tracking performance.

PG-Net~\cite{PG-Net} is a recently proposed part-based deep tracker; it decomposes the feature maps of the exemplar into spatial and channel kernels to perform pixel-to-global matching with the search image.
Similar to most part-based trackers, this method also equally treats all spatial kernels that represent a local part of the exemplar without considering their discriminability.
Unlike PG-Net, our \emph{SAOT} adopts a saliency mining mechanism to focus on the discriminative parts of the exemplar.
Besides, we explicitly model the interactions between the captured saliencies to effectively associate them, instead of directly combining the matching results of the parts by global matching as PG-Net does.
\vspace{-1.25mm}

\section{\mymodel}
\vspace{-1mm}
Given an exemplar image for the initial target and a search image in a tracking sequence, the goal of our \mymodel (\emph{SAOT}) is to learn robust correlation representations between them, which is able to effectively cope with various appearance variations of the target object during tracking, such as deformation and occlusion. To this end, our \emph{SAOT} first captures the local saliencies in the target object that are discriminative for tracking by the proposed Saliency Mining module, then models the associations between these saliencies to learn effective global correlation features between the target exemplar and the search image for precise tracking.

\vspace{-1mm}
\subsection{Overall Framework}
\vspace{-1mm}
Figure~\ref{Fig:Framework} illustrates the overall framework of the proposed \emph{SAOT}, consisting of two core modules: Saliency Mining module and Saliency-Association Modeling module. 

Taken as input an exemplar image and a search image in a tracking sequence, our \emph{SAOT} first employs a Siamese feature extractor to learn deep representations $\bm F_{x}\in \mathbb{R}^{h_x\times w_x\times c}$ and $\bm F_{s}\in \mathbb{R}^{h_s\times w_s\times c}$ in the same feature space for the target exemplar (cropped from the exemplar image according to the bounding box) and the search image, respectively. Herein we adopt widely used ResNet~\cite{ResNet} pre-trained on Imagenet~\cite{Imagenet} as the feature extractor due to its excellent performance of image feature learning.

The Saliency Mining module is designed to capture the local saliencies of the target exemplar which are discriminative for tracking. It calculates similarity maps to measure the pixel-to-pixel correspondences between $\bm F_{x}$ and $\bm F_{s}$, and selects local sharp maximum points as saliencies.
These captured saliencies correspond to the most discriminative regions of the exemplar, which can be easily localized with high confidence and accuracy.

The captured saliencies are then associated together by the Saliency-Association Modeling module of \emph{SAOT} to learn effective global correlation representations between the exemplar and the search image. The obtained correlation representations are expected to reflect the target state in the search image precisely by aggregating the distributions of all saliencies in the search image with the learned interactions between them. Finally, the target state is estimated by a classification head for confidence estimation and a regression head for predicting the bounding box of the target.

\vspace{-1mm}
\subsection{Saliency Mining}
\vspace{-1mm}
Typically, not all local regions of the target exemplar are easy to be tracked. Thus we design the Saliency Mining module to capture the saliencies corresponding to discriminative local regions of the target exemplar that can be easily localized in the search image. 

The proposed Saliency Mining module performs saliency mining in two steps: 1) constructs similarity maps for each pixel in the feature maps of the target exemplar $\bm F_{x}$ to achieve the distribution of matching score in the search image; 2) measures the saliency value of each pixel in $\bm F_{x}$ based on the obtained similarity maps to select saliencies. 

\begin{figure}[t]
\centering
    \includegraphics[width=0.905\columnwidth]{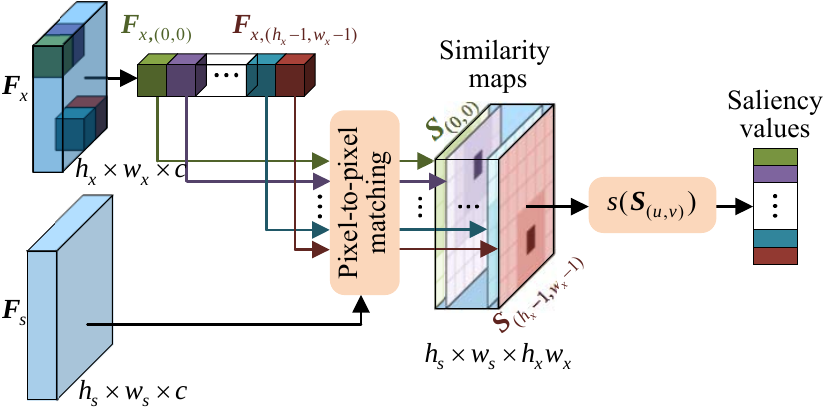}
    \caption{\textbf{Structure of the Saliency Mining module.} It first constructs the similarity maps by performing pixel-to-pixel matching between $\bm F_{x}$ and $\bm F_{s}$, and then computes the saliency value for every pixel in $\bm F_{x}$ based on the corresponding similarity map.}
\label{Fig:Matching}
\vspace{-4mm}
\end{figure}

\noindent\textbf{Construction of similarity maps.}
As shown in Figure~\ref{Fig:Matching}, the similarity map is constructed for each pixel in $\bm F_{x}$ by measuring the pixel-to-pixel matching degree between this pixel and the each pixel in $\bm F_{s}$. To be specific, the matching degree between the pixel located at $(u, v)$ in $\bm F_{x}$ and the pixel located at $(p, q)$ in $\bm F_{s}$ is formulated as:
\vspace{-1.5mm}
\begin{equation}
\label{Eq:similarity}
\bm S_{((u,v),(p,q))}=f(\bm F_{x,(u,v)},\bm F_{s,(q,p)}),
\vspace{-1.5mm}
\end{equation}
where $\bm F_{x,(u,v)}$ denotes the vectorial representations at location $(u,v)$ in $\bm F_{x}$ along the channel dimension, and similar denotation applies to $\bm F_{s,(p,q)}$. Herein $f$ refers to a kernel function for measuring similarity between two vectors. In our implementation, the cosine similarity operator is adopted for $f$, which is an efficient and effective distance metric. Hence, the similarity in Eq.~\ref{Eq:similarity} is calculated by:
\vspace{-1.5mm}
\begin{equation}
\label{Eq:cosine_similarity}
\bm S_{((u,v),(p,q))}=\frac{\bm F_{x,(u,v)}\cdot\bm F_{s,(q,p)}}{\|\bm F_{x,(u,v)}\|\|\bm F_{s,(p,q)}\|},
\vspace{-1.5mm}
\end{equation}
where $\cdot$ denotes the inner product operator.
The achieved similarities between the pixel at location $(u,v)$ in $\bm F_{x}$ and all pixels in $\bm F_{s}$ form a single-channel similarity map denoted as $\bm S_{(u,v)}\in \mathbb{R}^{h_s\times w_s}$.

\noindent\textbf{Saliency evaluation.}
For each pixel in the exemplar features $\bm F_{x}$, the maximum point in its similarity map is considered to be the matched position (with the largest confidence) for this pixel in the search image. We evaluate the saliency for this pixel based on the measurements of the peak distribution around the maximum point in the similarity map. Specifically, we consider two measurements: the intensity and the concentration of the peak distribution. 

\begin{figure}[t]
\centering
\includegraphics[width=1.3in]{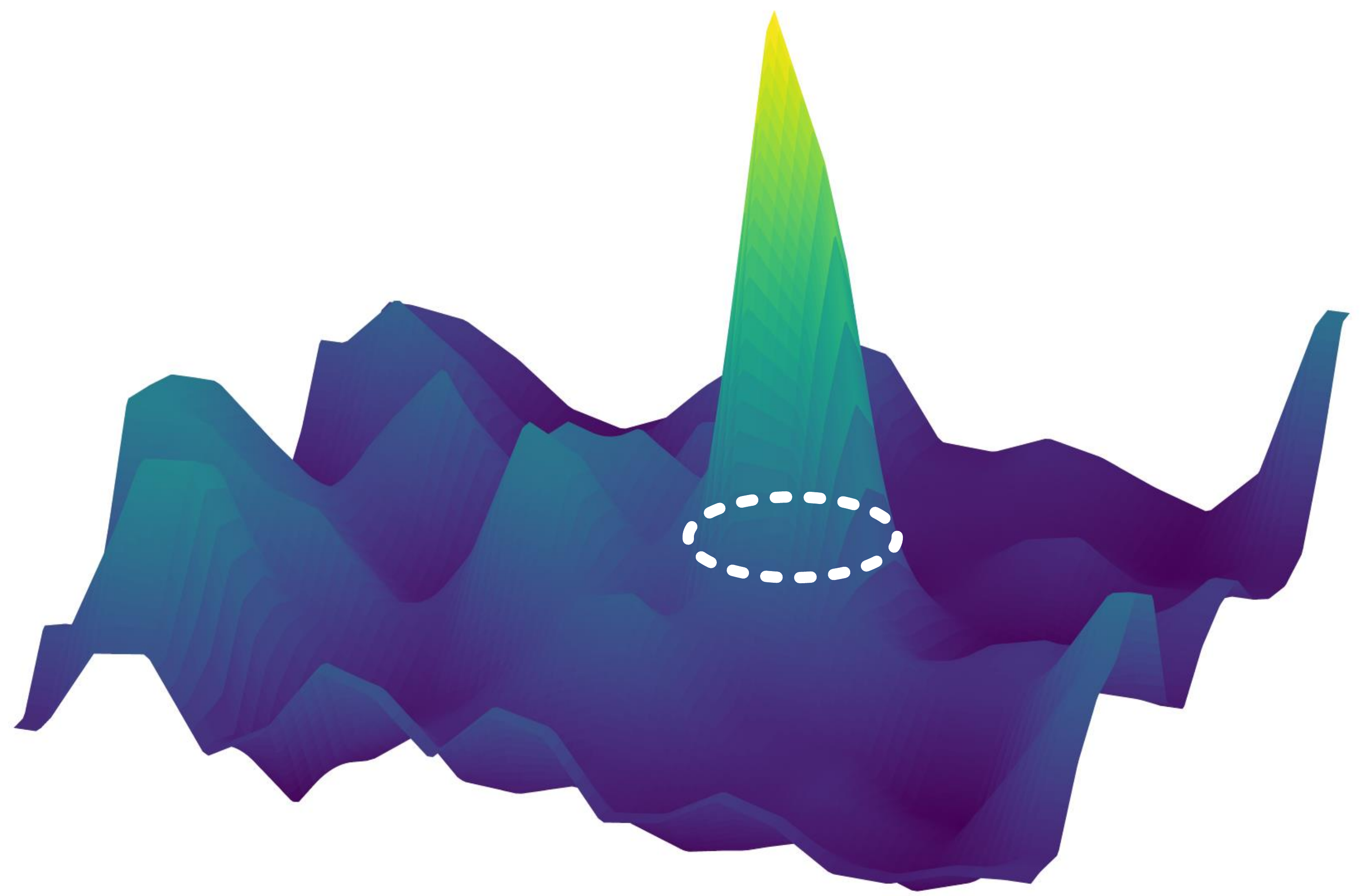}\hspace{0.135in}
\includegraphics[width=1.3in]{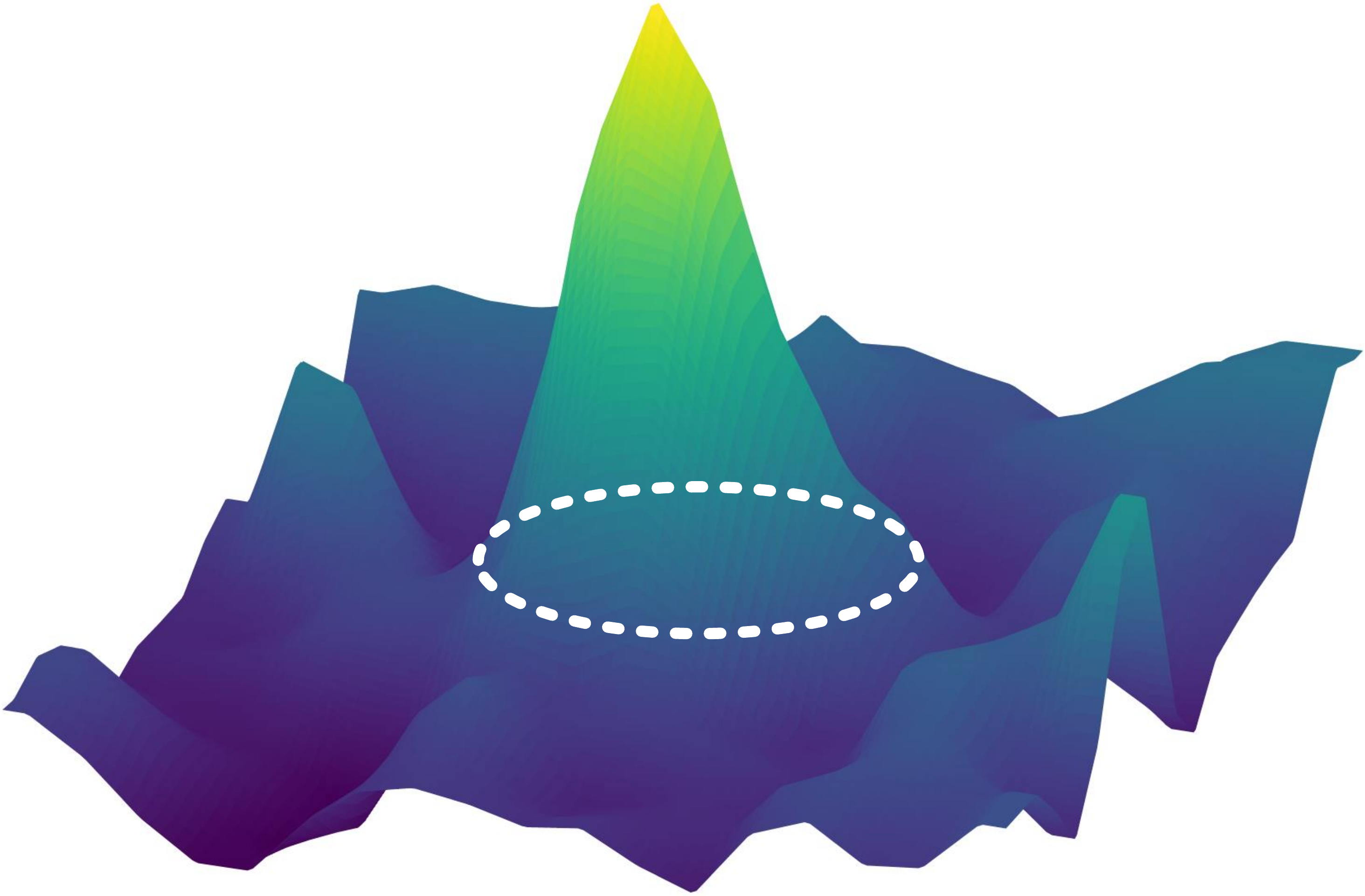}
\vspace{-0.5mm}
\caption{\textbf{Two similarity maps with different peak distributions.} The surface plots show the similarity values. Note that the main lobes, denoted by white dash circles, are of different sizes.}
\vspace{-4mm}
\label{Fig:Similarity_map}
\end{figure}

The intensity of a peak distribution is used to measure the relative strength of the maximum value compared to other values in the whole similarity map. A straightforward way to measure the intensity of a peak distribution is Peak-to-Sidelobe Ratio (PSR)~\cite{MOSSE} which is defined as:
\vspace{-1.5mm}
\begin{equation}
\label{Eq:psr}
{\rm PSR}(\bm S_{(u,v)}; \Phi) = \frac{{\rm max}(\bm S_{(u,v)})-\mu_{\Phi}(\bm S_{(u,v)})}{\sigma_{\Phi}(\bm S_{(u,v)})}.
\vspace{-1.5mm}
\end{equation}
Herein $\Phi$ denotes the sidelobe w.r.t.~to a peak distribution in the similarity map $\bm S_{(u,v)}$, which is defined as the region of $\bm S_{(u, v)}$ excluding the neighboring region around the maximum point (referred to as main lobe $\Psi$). Here main lobe and sidelobe are defined to roughly indicate the relevant and irrelevant regions to the peak distribution around the maximum point, respectively. $\mu_{\Phi}$ and $\sigma_{\Phi}$ are the mean value and standard deviation of $\bm S_{(u,v)}$ of the sidelobe, respectively. In the initial definition~\cite{MOSSE}, the size of main lobe $\Psi$ for arbitrary similarity maps is pre-defined as a fixed value. We argue that such a definition is unreasonable since the distribution characteristics of similarity maps are not taken into account. Figure~\ref{Fig:Similarity_map} shows two examples of similarity maps with different peak distributions around the maximum points, which apparently correspond to different sizes of the main lobes. Instead of fixing the size of main lobe, we define the boundary of main lobe $\Psi$ as the closest contour around the peak, whose height value is equal to the mean value of the similarity map. Consequently, the intensity $\gamma$ of a peak distribution in the similarity map $\bm S_{(u,v)}$ is defined as:
\vspace{-1.5mm}
\begin{equation}
    \begin{split}
        & \Psi \triangleq \text{region}(\bm S_{(u,v)}) \mid_{\text{contour}(\text{avg}(\bm S_{(u,v)}))}, \\
        & \gamma(\bm S_{(u,v)}) = {\rm PSR}(\bm S_{(u,v)}; \text{region}(\bm S_{(u,v)}) - \Psi),
    \end{split}
\vspace{-3mm}
\end{equation}
where $\text{avg}(\bm S_{(u,v)})$ is the mean value of the similarity map.

Another measurement we use for saliency evaluation is the concentration of the peak distribution, which is inversely proportional to the coverage area of the peak distribution around the maximum point. Thus, we measure the concentration $c$ of a peak distribution in the similarity map $\bm S_{(u,v)}$ by the reciprocal of the area of main lobe $A_\Psi(\bm S_{(u,v)})$:
\vspace{-2mm}
\begin{equation}
    c(\bm S_{(u,v)}) = A_\Psi^{-1}(\bm S_{(u,v)}).
\vspace{-1.5mm}
\end{equation}

Combining the defined intensity and the concentration, we evaluate the quality of a saliency $s( \bm S_{(u,v)})$ for the similarity map $\bm S_{(u,v)}$ by:
\vspace{-1.5mm}
\begin{equation}
    s(\bm S_{(u,v)}) = \gamma(\bm S_{(u,v)}) [c(\bm S_{(u,v)})]^\alpha,
\vspace{-1.5mm}
\end{equation}
where $\alpha$ is a hyper-parameter to balance between the effects of intensity and concentration.
The rationale behind this design is that the defined intensity and the concentration jointly reveal the sharpness of the peak distribution around the maximum point. A larger saliency value $s$ implies that the corresponding pixel in the feature maps of the exemplar $\bm F_x$ is more discriminative for tracking and easier to be localized in the feature maps of the search image $\bm F_s$.

Considering that the tracker should be encouraged to focus on tracking the central area of the target exemplar, a regularized term, which is a Gaussian mask, is added into the saliency evaluation metric $s$:
\vspace{-1.5mm}
\begin{equation}
    s(\bm S_{(u,v)}) = \gamma(\bm S_{(u,v)}) [c(\bm S_{(u,v)})]^\alpha + \lambda g_{\mu_g, \sigma_g}(u,v).
    \label{eqn:saliency_metric}
\vspace{-1.5mm}
\end{equation}
Herein $g_{\mu_g, \sigma_g}(u,v)$ is a Gaussian function aligned to the center of the exemplar feature $\bm F_x$, and $\lambda$ is a balance weight. During end-to-end training, the gradients can be back-propagated through the saliency evaluation metric, and we detail its back-propagation in the supplementary materials.

Based on the defined saliency evaluation metric in Eq.~\ref{eqn:saliency_metric}, we compute the saliency for each pixel in the feature maps of the exemplar $\bm F_x$ and select $K$ most salient pixels as the set of captured saliencies $P_{x}=\{\bm p_{x}^{k}\}_{k=1}^{K}$. The matched positions of these saliencies in the feature maps of the search image $\bm F_s$ composes the counterpart of the saliency set in the search image $P_{s}=\{\bm p_{s}^{k}\}_{k=1}^{K}$.

\vspace{-1mm}
\subsection{Saliency-Association Modeling}
\vspace{-1mm}
\label{Sec: Saliency-Association Modeling}
The captured saliencies, which are discriminative local parts of the target for tracking, are further associated together by the Saliency-Association Modeling module of \emph{SAOT} to learn effective global correlation representations between the exemplar and the search image. The obtained correlation representations are finally used for estimating the target state in the search image for tracking.

An intuitive way to associate the captured saliencies is to make connections between these local saliencies to form a global graph that is able to characterize the whole target. Following this way, the Saliency-Association Modeling module of our \emph{SAOT} performs saliency association in two steps: 1) constructs an effective graph among the captured saliencies to model the interactions between these saliencies; 2) aggregates the saliencies based on the constructed graph to learn global correlation representations between the exemplar and the search image.

\vspace{0.4mm}
\noindent\textbf{Construction of the saliency graph.}
We consider two types of information for node features when constructing the saliency graph: 1) the similarity maps $\bm S$ which contain the precise correspondence information from each local parts of the exemplar to the search image; 2) the feature maps of the search image $\bm F_s$. Two types of information, which have the equal size of feature maps ($h_s \times w_s$), are concatenated together in depth. Consequently, the resulting stacked feature maps (denoted as $\bm F_g$) can be considered as a graph which has total $h_s w_s$ regular nodes, while each node is represented by a vectorial feature whose dimension is $h_x w_x + c$. Note that the similarity maps are normalized by the corresponding saliency values before concatenation to emphasize more on the captured saliencies. Besides, the positions of the captured $K$ saliencies in the graph are indicated in $P_s$ obtained in the Saliency Mining module.

\begin{figure}[t]
\centering
    \includegraphics[width=0.775\columnwidth]{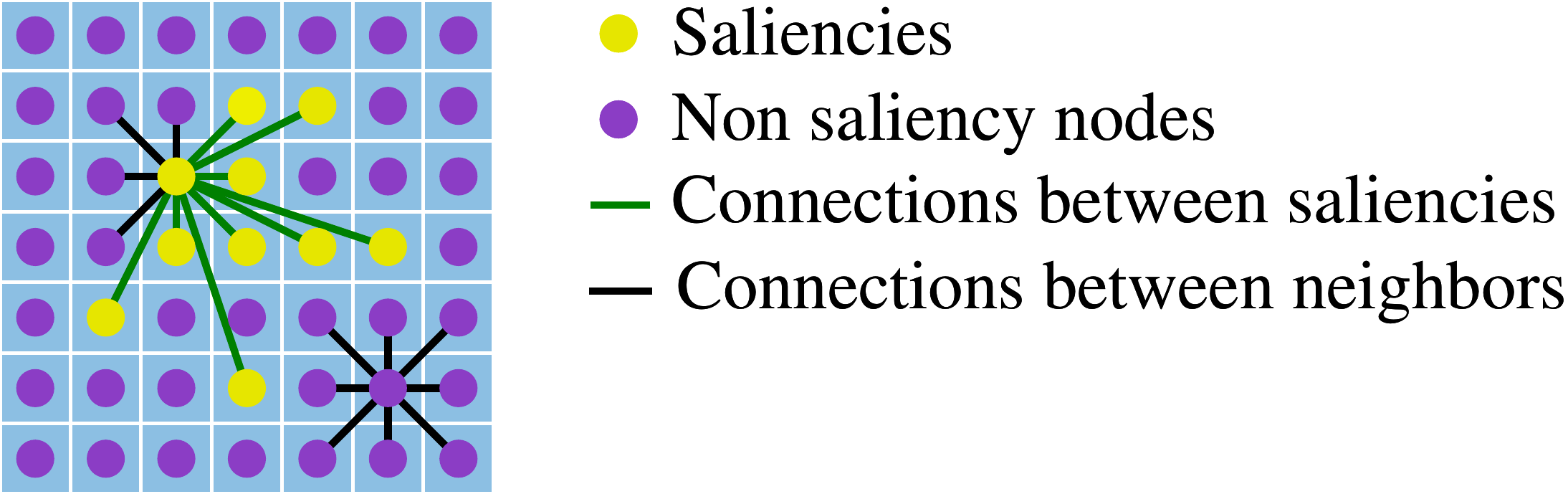}
    \caption{\textbf{Two kinds of connections considered for constructing the saliency graph.} The connections between saliencies are used to model the interactions between them, and those between neighbors are used for feature fusing between adjacent nodes.
    }
\label{Fig:Graph}
\vspace{-4mm}
\end{figure}

A key step of constructing the saliency graph is to model the interactions between nodes by connecting edges. Since we aim to associate the captured saliencies to achieve the effective global representation of the tracking target, we make pairwise edge-connections between $K$ saliencies specified in $P_s$. Besides, we also connect each node in $\bm F_g$ to its eight neighbors to perform neighboring information interactions for feature fusing between adjacent nodes. The resulting connection set including these two types of edges is denoted as $\mathcal{C}$, which is illustrated in Figure~\ref{Fig:Graph}.

To precisely model the interactions between the above specified connections, the edge weights are learned by the proposed Saliency-Association Modeling module rather than being fixed as binary values. Particularly, we customize a two-layer perception network to learn the edge weight for each connection specified before. Thus, the weighted adjacency matrix $\bm A \in \mathbb{R}^{N\times N}, N=h_s w_s$ for the saliency graph is modeled by:
\vspace{-1.5mm}
\begin{equation}
    \label{eq:adj}
    \resizebox{1.0\hsize}{!}{$\!\bm A_{ij}\!=\!\left\{
        \begin{array}{rcl}
        \!\!\!\sigma(\phi_{2}(\text{ReLU}(\phi_{1}(|\bm v_{i}\!-\!\bm v_{j}|)))), & &\!\!\!\!\!\!\!\!\!\! {if\ \text{edge} <\!i,j\!> \ \in \mathcal{C};}\\
        \!\!0, & &\!\!\!\!\!\!\!\!\!\! {else.}
        \end{array} \right.$}
\end{equation}
where $\bm v_i \in \mathbb{R}^{h_x w_x + c}$ and $\bm v_j\in \mathbb{R}^{h_x w_x + c}$ are features of two nodes in a connection. $\phi_{1}$ and $\phi_{2}$ denote the parameters of two fully connected layers, while $\sigma$ refers to Sigmoid function which transforms the edge weights to lie in $(0, 1)$. 

\noindent\textbf{Aggregation of the captured saliencies.}
The second step for saliency association modeling is to aggregate the saliency information according to the constructed saliency graph. There are multiple ways to perform graph aggregation. We opt for Graph Convolutional Networks (GCN)~\cite{SSC_GCN} for its effectiveness and convenience to be integrated into the whole model for end-to-end training. 

Specifically, we construct the two-layer GCN to perform saliency aggregation. Inspired by~Li et al.\cite{ASGCN}, we adopt the high-order polynomial of $\bm A$ to model multi-scale interactions between nodes. Formally, the $l$-th layer graph convolution is formulated as:
\vspace{-2mm}
\begin{equation}
\label{eq:GCN}
\bm X^{(l+1)}=\sigma_{l}(\sum \limits_{m=1}^{M}w_{m}\hat{\bm A}^{m}\bm X^{(l)}\bm \Theta_{(m)}^{(l)}),
\vspace{-2mm}
\end{equation}
where $m$ and $M$ are the polynomial order and total number of orders, respectively. 
Here, $w_{m}$ is a trainable weight for the order $m$.
$\hat{\bm A}=\tilde{\bm D}^{-\frac{1}{2}}\tilde{\bm A}\tilde{\bm D}^{\frac{1}{2}}$ is the normalized adjacency matrix~\cite{SSC_GCN}, where $\tilde{\bm A}=\bm A+\bm I$ and $\tilde{\bm D}$ is the diagonal degree matrix of $\tilde{\bm A}$.
$\bm X^{(l)}\in \mathbb{R}^{N\times d_{l}}$ and $\bm X^{(l+1)}\in \mathbb{R}^{N\times d_{l+1}}$ are the input and output features of all nodes at layer $l$ respectively, where $d_{l}$ and $d_{l+1}$ are the corresponding feature dimensions, and $d_0$ is equal to node feature dimension ($h_x w_x + c$).
$\bm \Theta_{(m)}^{(l)}\in \mathbb{R}^{d_{l}\times d_{l+1}}$ denotes the learnable parameter matrix at layer $l$ for $m$-th order. $\sigma_{l}$ is the activation function at layer $l$.

Through constructing the saliency graph and further performing saliency aggregation, the Saliency-Association Modeling module of \emph{SAOT} is able to learn global correlation representations between the target exemplar and the search image, which is further used for predicting the target state in the search image.

\vspace{-1.5mm}
\subsection{Tracking Framework}
\vspace{-1mm}
Our model can be readily integrated into various typical tracking frameworks. As shown in Figure~\ref{Fig:Framework}, we integrate our algorithm with a typical online learning tracker, namely online discriminative filter~\cite{DiMP}. The output global correlation representations by our algorithm is fed into a classification head for predicting the classification map and a regression head for predicting the bounding box of the target. In particular, the output $\bm p_{o}$ of the classification head is used to regularize the response map $\bm p_{r}$ produced by online discriminative filter via weighted element-wise summation to generate the final classification map $\bm p_{cls}$.
The predicted bounding box by the regression head, which is corresponding to the maximum classification score in $\bm p_{cls}$, is used as the final tracking result.
Both the classification and regression heads are designed following FCOS~\cite{FCOS}.

\vspace{0.4mm}
\noindent\textbf{End-to-end parameter learning.}
The whole model \emph{SAOT} is trained in an end-to-end manner.
Specifically, we employ IoU loss~\cite{GIoU} and binary cross-entropy (BCE) loss~\cite{Cross-entropy} to train the regression and classification heads respectively in an offline manner.
The online discriminative filter is trained following DiMP~\cite{DiMP}, whose offline training is performed jointly with the training of our \emph{SAOT}.
\vspace{-1.5mm}

\section{Experiments}
\vspace{-1mm}
\subsection{Experimental Setup}
\vspace{-1mm}
\noindent\textbf{Implementation details.}
We use the fused feature of \emph{conv-3} and \emph{conv-4} of ResNet~\cite{ResNet} as the Siamese representation for our \emph{SAOT}, where the fusion weights are computed according to SKNet~\cite{SKNet}.
The target exemplar is cropped from the feature maps of the exemplar image according to its bounding box and pooled by a PrPool~\cite{IoU-Net} layer to obtain its precise representation, whose size is set to $8\times 8$.
The search image is with an area $5^2$ times that of the target and resized to $288\times288$.
$\lambda$ and $\sigma_{g}$ in Eq.~\ref{eqn:saliency_metric} are set to 1 and 2, respectively.
$K$ is set to 48. $\bm p_{r}$ is set to 0.8.
We use the training splits of COCO~\cite{COCO}, GOT10k~\cite{Got-10k}, TrackingNet~\cite{TrackingNet}, and LaSOT~\cite{LaSOT} to train our model.
During training, the parameters in ResNet are frozen, while the other parameters are optimized using ADAM~\cite{Adam} with a learning rate decayed from $1\times 10^{-3}$ to $8\times 10^{-6}$ and a weight decay of $1 \times 10^{-4}$ except for those of the online discriminative filter, whose training settings are following DiMP~\cite{DiMP}.
Codes and raw results are available at \href{https://github.com/ZikunZhou/SAOT.git}{https://github.com/ZikunZhou/SAOT.git}.

\noindent\textbf{Datasets and metrics.}
We evaluate our algorithm on the OTB2015~\cite{OTB2015}, NFS30~\cite{NFS30}, LaSOT~\cite{LaSOT}, VOT2018~\cite{VOT2018}, and GOT10k~\cite{Got-10k} datasets.
Specifically, both OTB2015 and NFS30 consist of 100 sequences.
They use precision and success to measure tracking performance, and the area under the curve (AUC) of the success plot is used for ranking.
LaSOT is a large-scale dataset containing 1,400 sequences in total and 280 sequences in the testing set. It uses precision, normalized precision, and success as performance metrics.
VOT2018 contains 60 sequences and uses expected average overlap (EAO) to measure the overall performance of trackers.
GOT10k contains 10,000 and 180 sequences in the training and testing splits, respectively; it uses average overlap (AO) and success rate (SR) as performance metrics.

\begin{table}[t]
\begin{center}
\caption{\textbf{AUC and precision (Pre.) for six variants of our \emph{SAOT} on OTB2015 and NFS30.} The best scores are marked in \textbf{bold}.}
\label{Tab:Abla}
\vspace{1mm}
\scriptsize
\renewcommand\arraystretch{1}
\resizebox{0.84\linewidth}{!}{
\begin{tabular}{l|ccccc}
\toprule
\multirow{2}{*}{Variants}   & \multicolumn{2}{c}{OTB2015} & \phantom{a} & \multicolumn{2}{c}{NFS30} \\
& AUC & Pre. &  & AUC & Pre. \\
\midrule
Base model & 0.651 & 0.860 && 0.597 & 0.703 \\
PPFM         & 0.687 & 0.881 && 0.625 & 0.736 \\
PAM      & 0.701 & 0.909 && 0.641 & 0.761 \\
\textbf{SAOT (Ours)}     & \textbf{0.714} & \textbf{0.926} && \textbf{0.656} & \textbf{0.778} \\
\midrule
DW-Corr         & 0.691 & 0.884 && 0.617 & 0.712 \\
PG-Corr       & 0.693 & 0.896 && 0.619 & 0.711 \\
\bottomrule
\end{tabular}}
\end{center}
\vspace{-8mm}
\end{table}

\vspace{-1.5mm}
\subsection{Ablation Study}
\vspace{-1mm}
\label{Seq:Ablation_Study}
To investigate the effectiveness of each proposed component, we perform ablation studies with six variants of \emph{SAOT}:

\noindent1)~\textbf{Base model}, which only contains the feature extractor, the classification and regression heads, and the online filter of \emph{SAOT}. Herein, the classification and regression heads are constructed on the feature maps of the search image $\bm F_{s}$. 

\noindent2)~\textbf{PPFM}, which computes the similarity maps $\bm S$ between the exemplar and the search image by Pixel-to-Pixel Feature Matching to improve the base model.
It uses a two-layer CNN to adjust the stacked feature maps of $\bm S$ and $\bm F_{s}$ to generate the correlation representation, on which the classification and regression heads are constructed.

\noindent3)~\textbf{PAM}, which associates all local parts to generate the correlation representation by viewing all the local parts equally as saliencies, i.e., no saliencies are captured. We denote such a model as Part-Association Model.

\noindent4)~\textbf{\emph{SAOT}}, our intact model which associates saliencies instead of all local parts as PAM does.

\noindent5)~\textbf{DW-Corr}, which employs the Depth-Wise cross Correlation~\cite{SiamRPN++} in our framework to replace the saliency mining and saliency-association modeling modules.

\noindent6)~\textbf{PG-Corr}, which employs the Pixel to Global cross Correlation~\cite{PG-Net} in our framework.

Table~\ref{Tab:Abla} presents the experimental results of these variants on the OTB2015~\cite{OTB2015} and NFS30~\cite{NFS30} benchmarks.

\noindent\textbf{Effect of the constructed similarity maps.} 
The performance gaps between base model and PPFM clearly demonstrate the benefits of constructing similarity maps in the feature space to model the fine-grained similarity between the exemplar and the search image.

\noindent\textbf{Effect of association modeling.}
Compared with PPFM, PAM achieves performance gains of 1.4\% and 1.6\% in AUC on OTB2015 and NFS30, respectively.
These results validate the benefits of associating the matched local parts by modeling the pairwise interactions between them, which generates a more robust correlation representation.

\begin{figure}[t]
\centering
\includegraphics[width=1.5in]{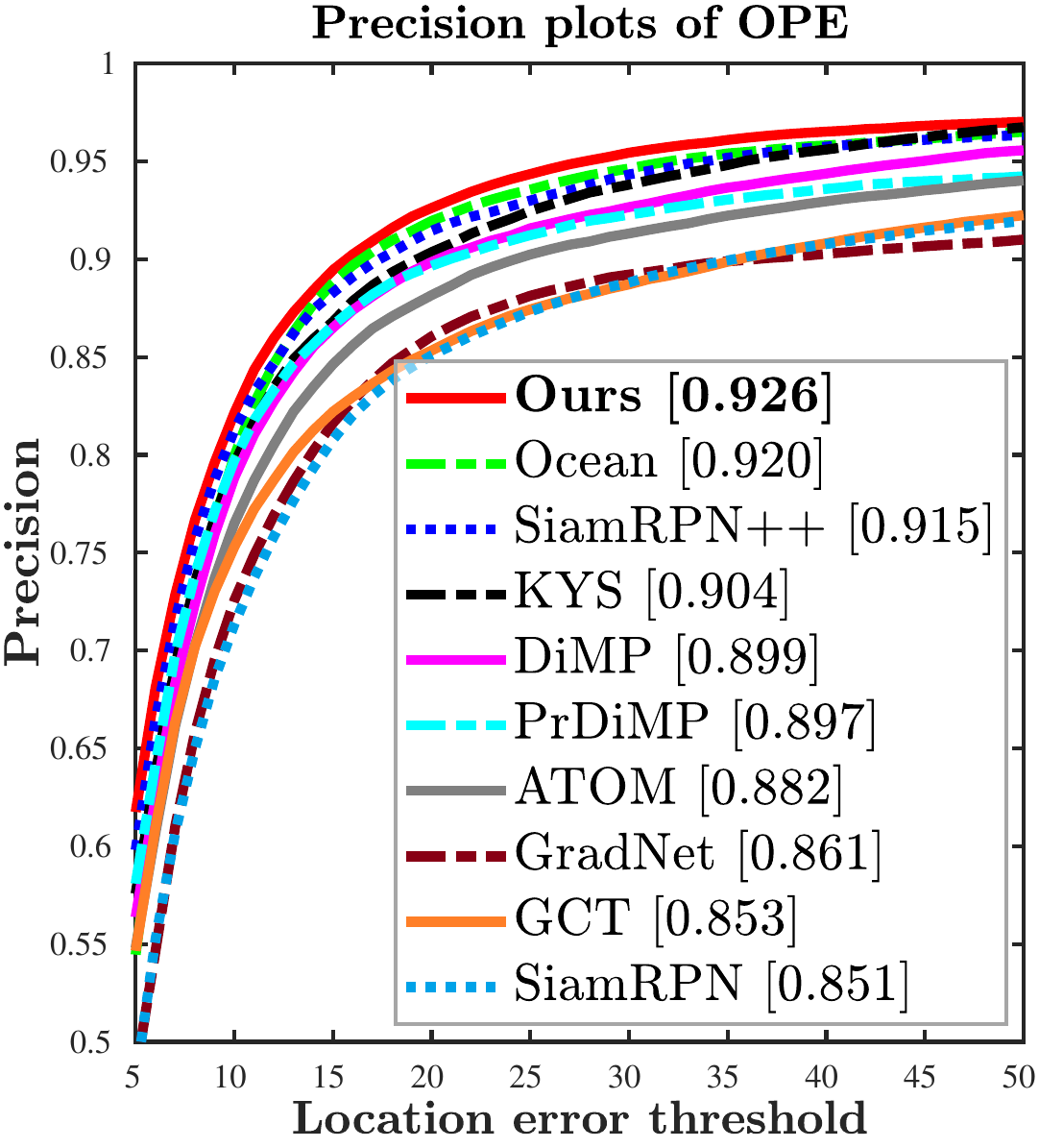}
\includegraphics[width=1.5in]{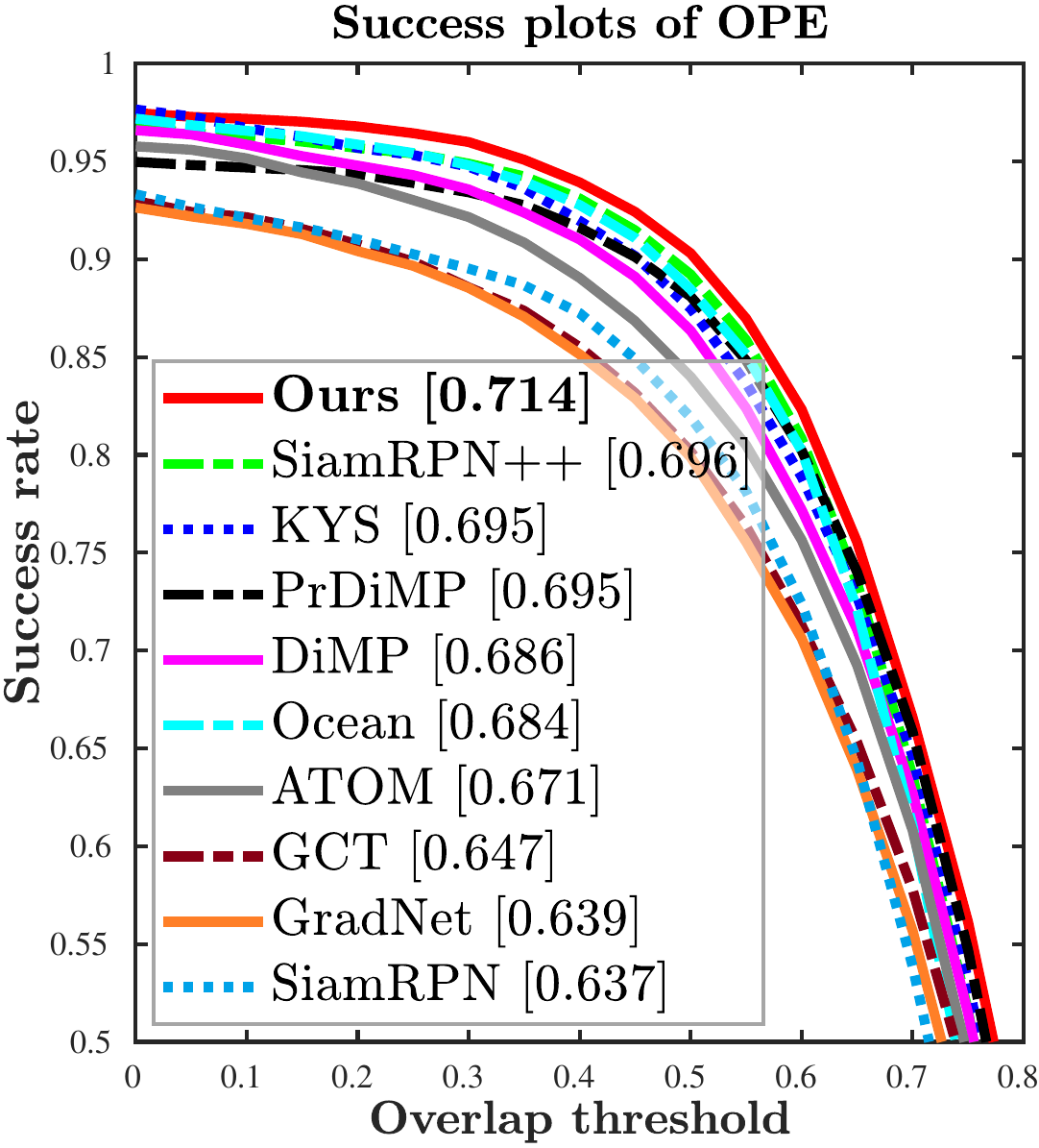}
\vspace{-0.75mm}
\caption{\textbf{Precision and success plots of different tracking methods on OTB2015.}}
\vspace{-1.mm}
\label{Fig:OTB2015}
\end{figure}

\begin{figure}[t]
\centering
\includegraphics[width=1.5in]{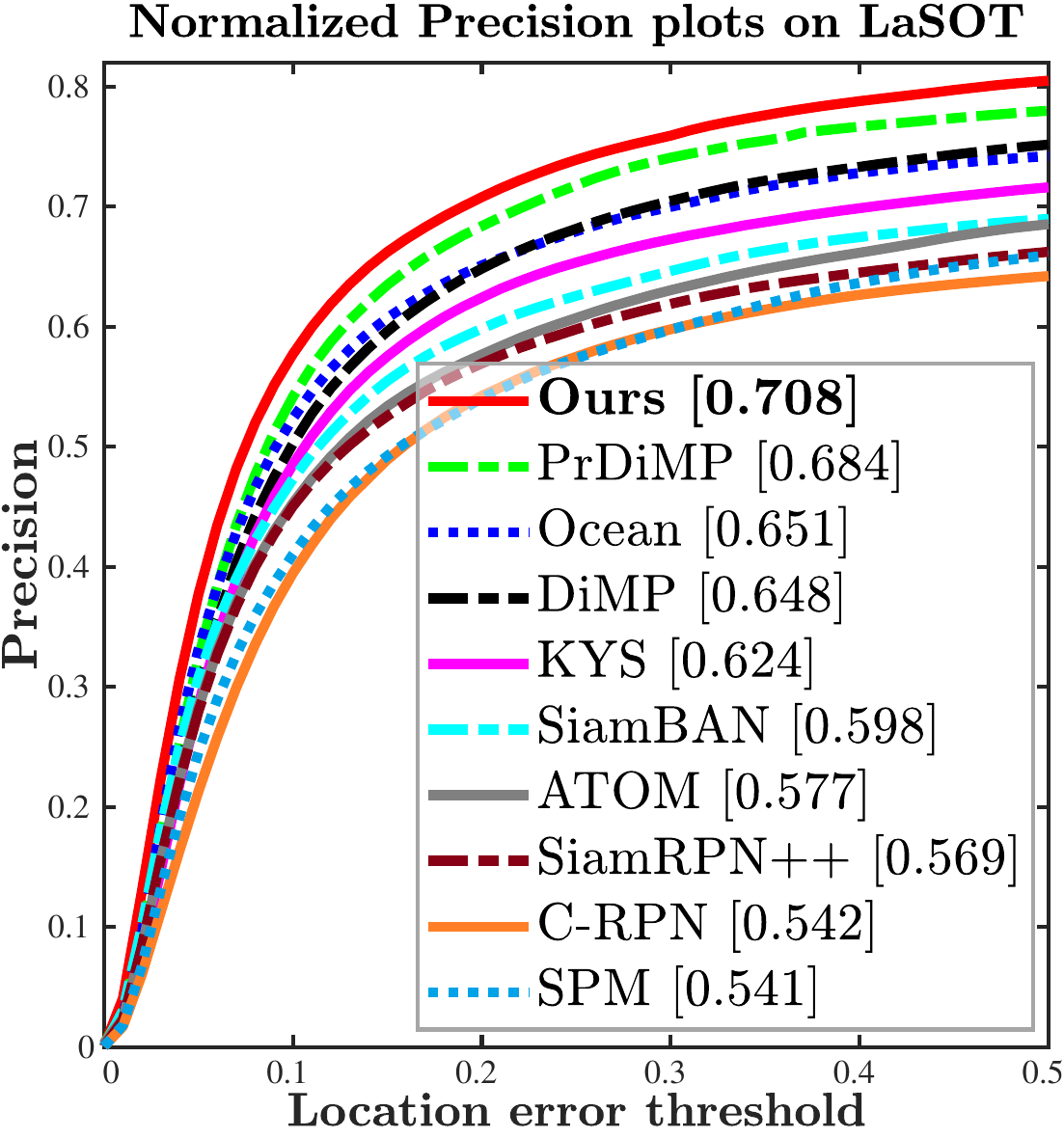}
\includegraphics[width=1.5in]{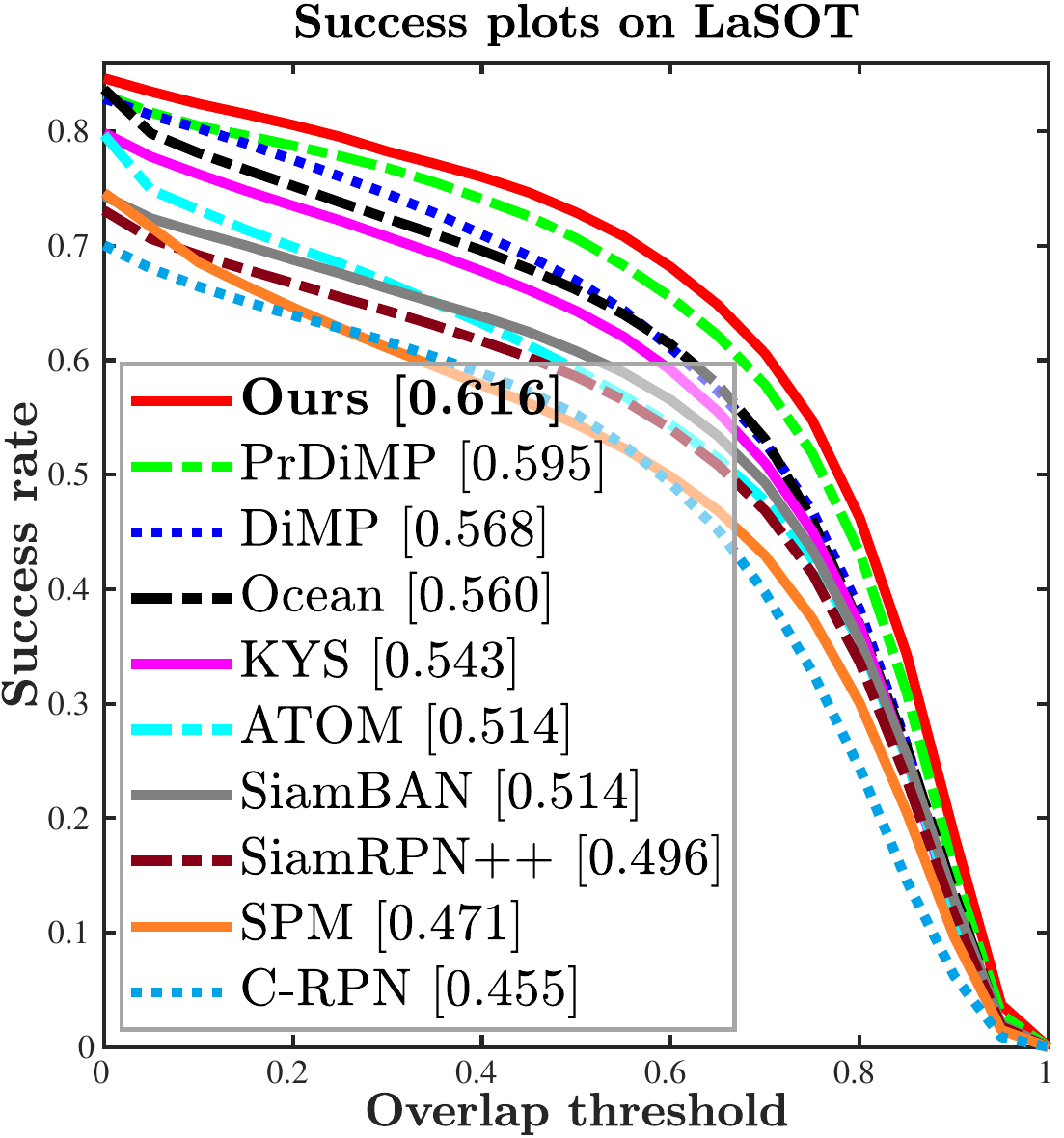}
\vspace{-0.75mm}
\caption{\textbf{Normalized precision and success plots of different tracking methods on the test set of LaSOT.}}
\vspace{-4.8mm}
\label{Fig:LaSOT}
\end{figure}

\noindent\textbf{Effect of the saliency mining mechanism.}
The comparison between PAM and our \emph{SAOT} manifests the effectiveness of the proposed saliency mining mechanism, which further improves tracking performance by 1.3\% and 1.5\% in AUC on OTB2015 and NFS30, respectively.
This mechanism successfully enables the tracker to focus on local saliencies of the target that are discriminative for tracking.

\noindent\textbf{Comparison between different correlation calculating methods.}
The performance of DW-Corr and PG-Corr decreases by 2.3\%/2.1\% and 3.9\%/3.7\% in AUC on OTB2015 and NFS30, respectively, compared with our \emph{SAOT}.
It demonstrates the superiority of the correlation representation learned by mining saliencies and associating them.

\begin{figure}[t]
\begin{center}
\includegraphics[width=0.925\columnwidth]{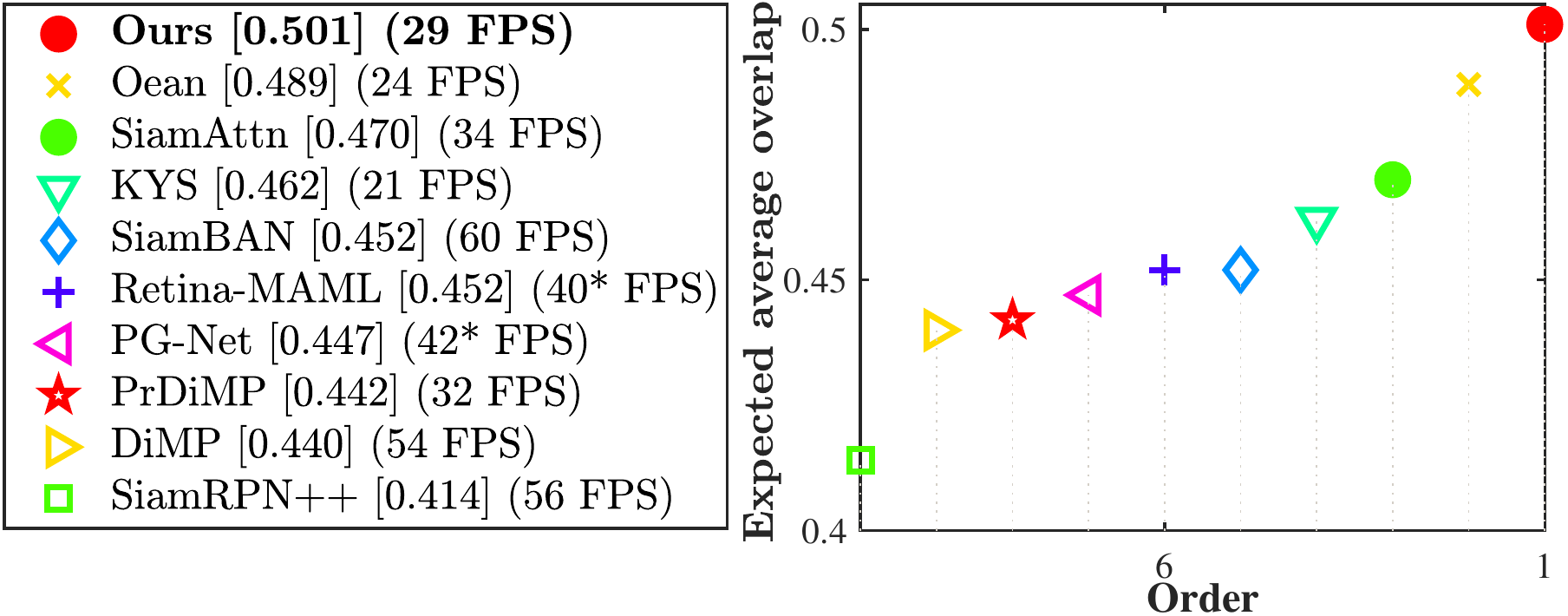}
\vspace{-0.5mm}
\caption{\textbf{Expected average overlap and average running speed of different trackers on VOT2018.} The notation * denotes the speed is reported by the authors as the code is not available.}
\vspace{-4.5mm}
\label{Fig:VOT2018}
\end{center}
\end{figure}

\vspace{-1mm}
\subsection{Comparison with State-of-the-art Trackers}
\vspace{-1mm}
Herein we compare our \emph{SAOT} with 17 representative state-of-the-art methods on five benchmarks, including OTB2015, NFS30, LaSOT, VOT2018, and GOT10k.
The methods involved in the comparison include 16 holistic-strategy trackers (KYS~\cite{KYS}, Ocean~\cite{OCEAN}, SiamBAN~\cite{SiamBAN}, SiamAttn~\cite{SiamAttn}, PrDiMP~\cite{PrDiMP}, Retina-MAML~\cite{MAML}, DiMP~\cite{DiMP}, GradNet~\cite{Gradnet}, ATOM~\cite{ATOM}, SiamRPN++~\cite{SiamRPN++}, C-RPN~\cite{C-RPN}, GCT~\cite{GCT}, SPM~\cite{SPM-Tracker}, DaSiamRPN~\cite{DaSiamRPN}, SiamRPN~\cite{SiamRPN}, and DSiam~\cite{DSIAM}) and one part-based tracker (PG-Net~\cite{PG-Net}). We discuss the experimetal results per dataset below.

\noindent\textbf{OTB2015}.
Figure~\ref{Fig:OTB2015} illustrates the precision and success plots on OTB2015.
Our algorithm achieves the best AUC score of 0.714 and the best precision score of 0.926.
Note that DiMP~\cite{DiMP}, Ocean~\cite{OCEAN}, and our \emph{SAOT} are all built based on the same online discriminative filter.
The difference is that DiMP and Ocean perform tracking in the holistic tracking strategy while our method adopts the part-based strategy. Our method outperforms these two methods by a large margin (2.8\% and 3.0\% in AUC, respectively), which demonstrates the effectiveness of the proposed method.

\newcommand{\tabincell}[2]{\begin{tabular}{@{}#1@{}}#2\end{tabular}}
\begin{table}[t]
\setlength\tabcolsep{2.5pt}
\begin{center}
\caption{\textbf{AUC of different tracking methods on NFS30.}}
\label{Tab:NFS30}
\vspace{0.5mm}
\scriptsize
\renewcommand\arraystretch{1}
\resizebox{1.0\linewidth}{!}{
\begin{tabular}{l|ccccccccc}
\toprule
           &\tabincell{c}{DaSiam\\RPN~\cite{DaSiamRPN}} & \tabincell{c}{SiamRPN\\++~\cite{SiamRPN++}}& 
           \tabincell{c}{ATOM\\~\cite{ATOM}} & \tabincell{c}{SiamBAN\\~\cite{SiamBAN}} &  \tabincell{c}{DiMP\\~\cite{DiMP}} &
           \tabincell{c}{KYS\\~\cite{KYS}} &
           \tabincell{c}{PrDiMP\\~\cite{PrDiMP}} & \textbf{Ours} \\
\midrule
AUC~   & 0.395 & 0.503 & 0.584 & 0.594 & 0.619 & 0.634 & 0.635 & \textbf{0.656}   \\
\bottomrule
\end{tabular}}
\vspace{-4mm}
\end{center}
\end{table}

\begin{table}[t]
\setlength\tabcolsep{2.5pt}
\begin{center}
\caption{\textbf{AO and SR of different trackers on GOT10k.}}
\label{Tab:GOT10k}
\vspace{-2mm}
\scriptsize
\renewcommand\arraystretch{1}
\resizebox{1.0\linewidth}{!}{
\begin{tabular}{l|ccccccccc}
\toprule
           & \tabincell{c}{DSiam\\~\cite{DSIAM}} &
           \tabincell{c}{SiamRPN\\++~\cite{SiamRPN++}} &
           \tabincell{c}{ATOM\\~\cite{ATOM}} & 
           \tabincell{c}{Ocean\\~\cite{OCEAN}} &   \tabincell{c}{DiMP\\~\cite{DiMP}} &
           \tabincell{c}{PrDiMP\\~\cite{PrDiMP}} & \tabincell{c}{KYS\\~\cite{KYS}} & \textbf{Ours} \\
\midrule
AO~        & 0.417 & 0.518 & 0.556 & 0.611 & 0.611 & 0.634 & 0.636 & \textbf{0.640} \\
SR$_{0.5}$~& 0.461 & 0.618 & 0.634 & 0.721 & 0.717 & 0.738 & -- & \textbf{0.749} \\
\bottomrule
\end{tabular}}
\vspace{-8mm}
\end{center}
\end{table}

\begin{figure*}[t]
\centering
    \includegraphics[width=0.88\textwidth]{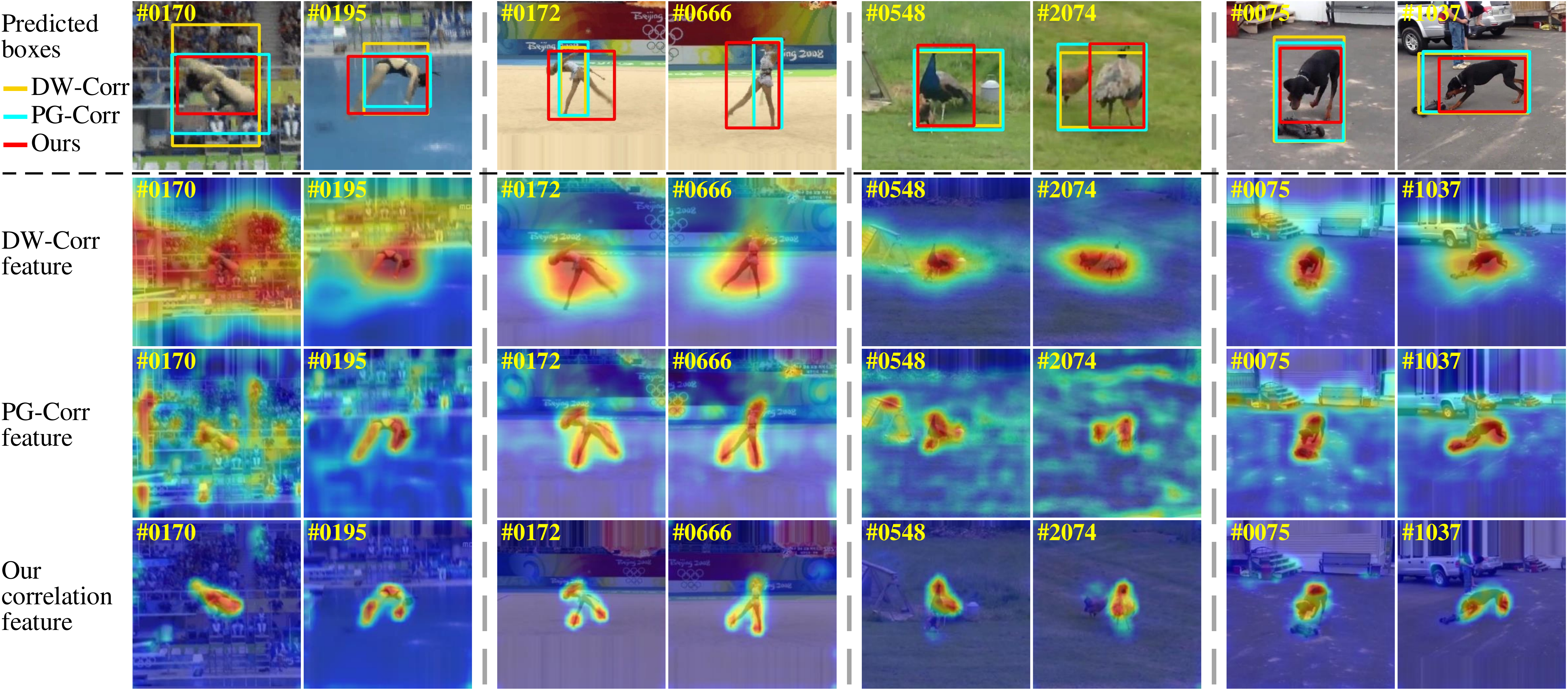}
\vspace{-0.5mm}   
\caption{\textbf{Qualitative comparison between our \emph{SAOT}, DW-Corr, and PG-Corr on four challenging tracking sequences (left two with deformation and the other two with distractors).} Our \emph{SAOT} is able to learn more precise correlation features than those generated by DW-Corr and PG-Corr. Consequently, our \emph{SAOT} predicts more precise bounding boxes than the other two methods.}
\label{Fig:Corr_feat_display}
\vspace{-4mm}
\end{figure*}

\noindent\textbf{NFS30}.
Table~\ref{Tab:NFS30} reports the AUC scores on NFS30.
While PrDiMP~\cite{PrDiMP} and KYS~\cite{KYS} perform well on this dataset with AUC scores of 0.635 and 0.634, respectively, the proposed \emph{SAOT}, achieving the best AUC score of 0.656, further improves tracking performance by 2.1\% and 2.2\% over these two trackers, respectively.

\noindent\textbf{LaSOT}.
We follow protocol II~\cite{LaSOT} to evaluate the proposed \emph{SAOT} on the test set of LaSOT.
Figure~\ref{Fig:LaSOT} shows the normalized precision and success plots.
Our \emph{SAOT} achieves the best performance in both AUC and normalized precision.
Compared to Ocean~\cite{OCEAN} and DiMP~\cite{DiMP}, our method achieves remarkable performance gains of 5.6\%/4.8\% and 5.7\%/6.0\% in AUC and normalized precision, respectively.

\noindent\textbf{VOT2018}.
Figure~\ref{Fig:VOT2018} presents the EAO scores of different trackers on VOT2018.
Although Ocean~\cite{OCEAN} obtains an impressive EAO score of 0.489, our method further improves the EAO score by 1.2\%.
Besides, compared with the state-of-the-art online discriminative filter-based methods KYS~\cite{KYS} and PrDiMP~\cite{PrDiMP}, our \emph{SAOT} achieves substantial performance gains of 3.9\% and 5.9\% in EAO, respectively.
We also report the average running speeds of different trackers in Figure~\ref{Fig:VOT2018}, which are tested using the same PC with an RTX2080 GPU on VOT2018 without reset. Our \emph{SAOT} runs at 29 FPS, achieving real-time performance.

\noindent\textbf{GOT10k}.
We follow the defined protocol~\cite{Got-10k} to train our \emph{SAOT} for evaluating it on GOT10k.
Table~\ref{Tab:GOT10k} reports the AO and SR scores on the test set of GOT10k.
Compared with Ocean~\cite{OCEAN} and DiMP~\cite{DiMP}, the proposed method achieves performance gains of 2.9\%/2.9\% in AO and 2.8\%/3.2\% in SR$_{0.5}$, respectively.
In addition, our algorithm performs favorably against PrDiMP~\cite{PrDiMP} and KYS~\cite{KYS}.
These experimental results on GOT10k, whose training and testing sets do not share the object class, validates the generalization ability of our approach across different object classes.

\vspace{-1.25mm}
\subsection{Qualitative Study}
\vspace{-1mm}
To obtain more insights into our method, we visualize the correlation representations and the saliency values.

\begin{figure}[t]
\centering
    \includegraphics[width=0.92\columnwidth]{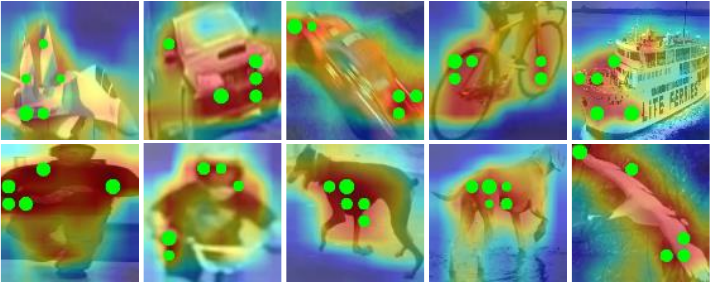}
\vspace{-0.5mm}    
\caption{\textbf{Saliency maps of 10 target exemplars.} The saliency maps are obtained by visualizing the saliency scores calculated by Eq.~\ref{eqn:saliency_metric} for the whole feature maps of each exemplar. Top-5 most salient local regions are indicated in green dots for each exemplar.}
\label{Fig:Saliency}
\vspace{-5mm}
\end{figure}

\noindent\textbf{Visualization of correlation representations.}
We visually compare our \emph{SAOT} and the two other variants DW-Corr and PG-Corr.
Figure~\ref{Fig:Corr_feat_display} shows the correlation features and the bounding boxes on two challenging sequences with deformation (left) and two with distractors (right).
Our model predicts more precise correlation features and bounding boxes than the other two methods, which implies its better capability to handle deformation and distractors as the captured saliencies are robust to deformation and distractors.

\noindent\textbf{Visualization of saliency maps.}
Figure~\ref{Fig:Saliency} illustrates the saliency maps of ten target exemplars.
We observe that the proposed saliency evaluation metric assigns high saliency values to local regions that are discriminative for tracking.
\vspace{-2mm}

\section{Conclusion}
\vspace{-1.5mm}
In this work, we have presented the Saliency-Associated Object Tracker (\emph{SAOT}), which first deals with the discriminative local saliencies and then associates them to achieve the global solution.
Specifically, our \emph{SAOT} employs the proposed Saliency Mining module to capture the saliencies of the target object, which are robust to target deformation and distractors.
Further, we propose a Saliency-Association Modeling module to associate the captured saliencies by modeling the interactions between them, learning a precision correlation representation for reflecting the target state.
The proposed method achieves favorable performance against state-of-the-art trackers on five datasets.
\vspace{-2mm}

\section*{Acknowledgments}
\vspace{-2mm}
This work was supported by the National Natural Science Foundation of China (U2013210, 62006060, and 62002241), the Special Research project on COVID-19 Prevention and Control of Guangdong Province (2020KZDZDX1227), the Shenzhen Research Council (JCYJ20170413104556946), and the Shenzhen Stable Support Plan Fund for Universities (GXWD20201230155427003-20200824125730001).

{\small
\bibliographystyle{ieee_fullname}
\bibliography{tracking}
}
\end{document}